\documentclass[runningheads]{llncs}

\usepackage[T1]{fontenc}

\usepackage{amsmath,amssymb}
\usepackage{graphicx}
\graphicspath{{./img}}

\usepackage[utf8]{inputenc}

\usepackage{hyperref}
\usepackage{color}

\usepackage{booktabs}

\usepackage[table]{xcolor}
\definecolor{orange_task}{HTML}{FF8C00} % [38;5;208m
\definecolor{okcyan}{HTML}{17A2B8}   % [96m
\definecolor{purple_task}{HTML}{6F42C1} % [95m
\definecolor{tab_blue}{HTML}{1F77B4}
\definecolor{tab_red}{HTML}{D62728}
\definecolor{tab_gray}{HTML}{7F7F7F}
\definecolor{tab_purple}{HTML}{9467bd}
\definecolor{tab_pink}{HTML}{e377c2}
\definecolor{tab_orange}{HTML}{ff7f0e}
\usepackage{xspace}
\newcommand{\randct}{\textcolor{tab_gray}{\textbf{\textsf{R}}}\xspace}
\newcommand{\dcct}{\textcolor{tab_blue}{\textbf{\textsf{C}}}\xspace}
\newcommand{\sdcct}{\textcolor{tab_red}{\textbf{\textsf{S}}}\xspace}

\DeclareMathOperator{\dataset}{\mathcal{D}}

\begin{document}
\title{Distance generalization in transformers: why bother with
       positional encoding?}
\titlerunning{Distance generalization}
% If the paper title is too long for the running head, you can set
% an abbreviated paper title here
%
\author{Daniel Henrik Nevermann
% \footnote{\orcidID{0000-0002-4607-5142}} 
\and
Claudius Gros
% \footnote{\orcidID{0000-0002-2126-0843}}
}
\authorrunning{D. H. Nevermann et al.}
% First names are abbreviated in the running head.
% If there are more than two authors, 'et al.' is used.
%
\institute{Institute for Theoretical Physics, 
Goethe University Frankfurt, Germany\\
\email{\{nevermann,gros\}@itp.uni-frankfurt.de}}
\maketitle              % typeset the header of the contribution

%\verb|\textwidth:| \the\textwidth

\begin{abstract}
    Out-of-distribution length generalization, namely to extrapolate a task from short 
    to longer context, has been studied intensively for transformers. Here we focus on \emph{distance generalization}, which probes performance when inter-token distances are changed between training and inference, while keeping a fixed context length. We construct two synthetic delay copy tasks, both involving finite distances between source and recall, where tokens are copied either fully or selectively, and test models on delays unseen during training. We address three questions: (A) Do positional encoding schemes such as RoPE and ALiBi improve distance resolution relative to no
    positional encoding (NoPE)? (B) How does data diversity, the number of inter-token distances seen in training, affect performance? (C) When is distance transfer learning positive or negative? We present a thorough investigation, finding that it is paramount to improve our understanding of the underlying mechanisms.
 
\end{abstract}

\section{Introduction}
A core challenge for transformer-based large language models is their ability to
generalize to out-of-distribution examples. It is well documented that
extracting latent patterns in the training data enables language models to
solve unseen problems at inference time \cite{power2022grokking,li2022emergent}.
Specifically, \emph{length generalization}, that is the ability
of a model to generalize tasks from short to longer context lengths, remains an
area of active investigation \cite{izzo2025quantitative,cai2025extrapolation}.

\begin{figure}[t]
    \centering
    \includegraphics{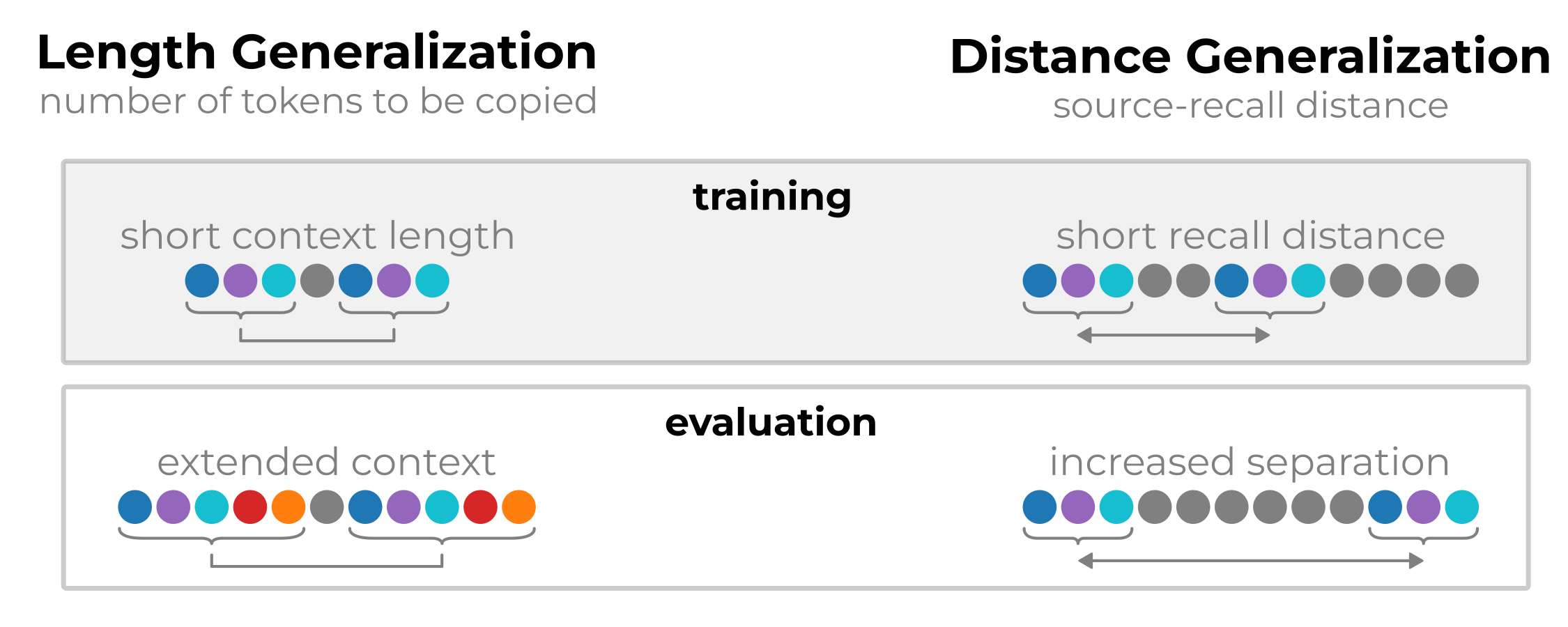}
    \caption{{\sl Comparison of length generalization and distance generalization.}
    Illustration for a basic copy task (see e.g.\ \cite{kazemnejad2023impact}). 
    For length generalization, the model is evaluated to copy sequences longer than the
    training sequences, resulting in longer context lengths at inference time.
    For distance generalization the context length remains fixed. Instead, 
    part of the sequence is copied with delay, where the distance between source 
    and recall (the delay) is increased or reduced at inference time.}
    \label{fig:fig1}
\end{figure}

In this work we focus on a related type of generalization, \emph{distance
generalization}, which is illustrated in Fig.~\ref{fig:fig1}.
Unlike length generalization, for which a model's performance beyond its 
training context is evaluated, we constrain the context length and
examine instead the model's ability to generalize across varying 
inter-token distances within an established context window.
Importantly, the complimentary investigation of distance and length 
generalization allows to isolate distinct failure modes,
respectively ``\emph{can't handle unseen positions}" versus
``\emph{can't handle unseen token dependencies}".

%%%%%%%%%%%%%%%%%%%%%%%%%%%%%%%%%%%%%%%%%%%%%%%%
%%%%%%%%%%%%%%%%%%%%%%%%%%%%%%%%%%%%%%%%%%%%%%%%

Following an increasing body of work on length generalization, 
we consider algorithmic tasks in our assessment of distance 
generalization. More precisely, we derive our datasets from 
a task switching framework that combines multiple different 
tasks within a single input sequence~\cite{gros2025small}, 
which means that active tasks are recurrently switched. We use task 
switching for two delay copy tasks, a full delay copy and a
selective delay copy, tailored to study
distance generalization. In Fig.~\ref{fig:fig1} we compare
a typical copy task used in length generalization studies with
the corresponding application to distance generalization.

\begin{figure}[t]
    \centering
    \includegraphics[width=\textwidth]{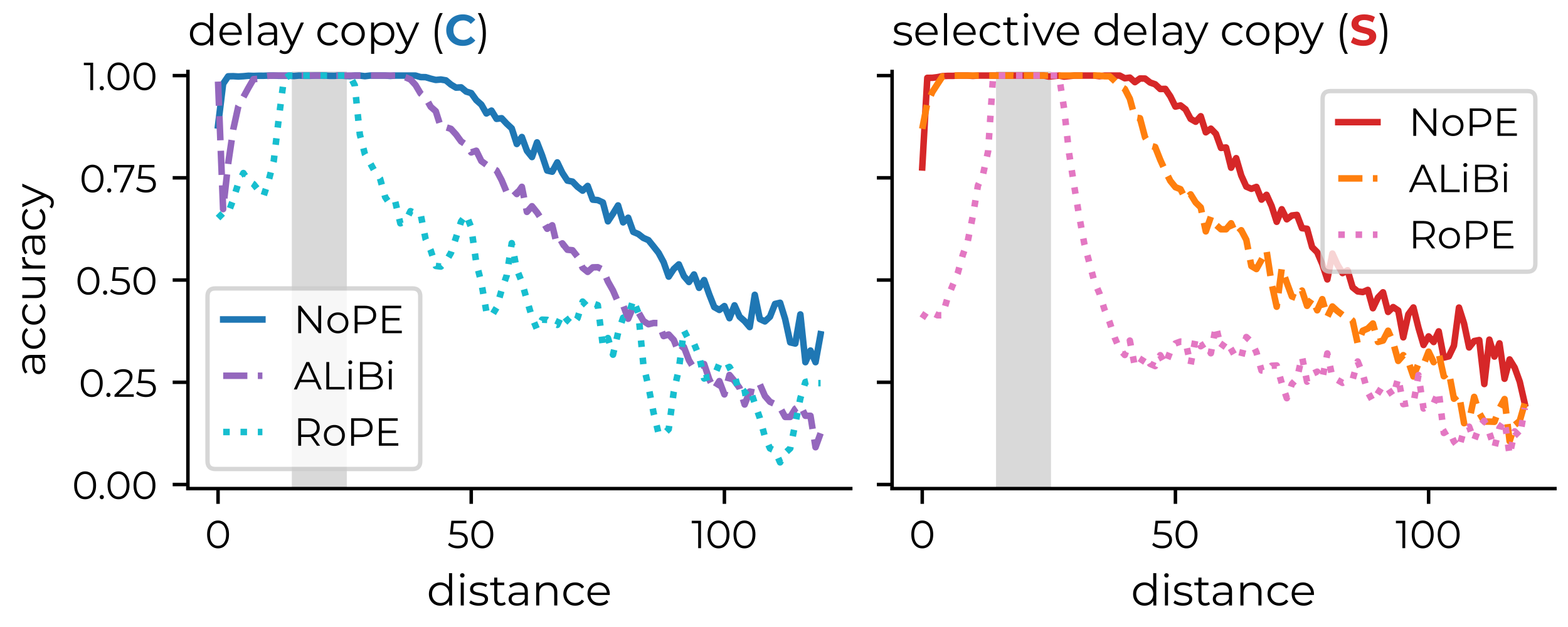}
    % \begin{itemize}
    %     \item for all PE ZL / ZS (n times for statistic)
    % \end{itemize}
    \caption{{\sl Distance generalization is strongly affected by positional
    encoding.} We compare two explicit positional encoding schemes,
    ALiBi and RoPE, with NoPE (no positional encoding), where NoPE relies soly
    on causality, viz causal attention for inferring relative distances.
    The evaluation accuracy is plotted as a function of the distance between
    source and recall, compare Fig.~\ref{fig:fig1}. Either all $m=10$ source
    tokens are to copied (left), the \dcct task, or only the even component (right),
    denoted selective copy task \sdcct.
    For details see Sec.~\ref{sec:experimental-setup}.
    During training, the model sees delay distances between 15 and 25
   (gray shaded region).
    }
    \label{fig:pes_by_task}
\end{figure}

Our results demonstrate that evaluating distance generalization 
is essential for a full understanding of how transformers 
generalize across positions. We identify different drivers 
and inhibitors of distance generalization.
As a first step we study the impact of positional
encoding on distance generalization. As illustrated
in Fig.~\ref{fig:pes_by_task}, one observes a strong impact.
Somewhat paradoxically, we find that the absence
of an explicit positional encoding scheme (NoPE) leads to the best
generalization capabilities, in agreement with findings in 
\cite{kazemnejad2023impact} on length generalization. Details
are discussed in Sec.~\ref{sec:results}.

We investigate furthermore how distance generalization is 
impacted by the diversity of distances presented in the training 
dataset. We find that training with a larger set of inter-token 
distances does increase generalization capabilities in absolute
terms, somewhat as expected, but we also report strongly diminishing 
returns when performance is evaluated in relative terms.
See Fig.~\ref{fig:dist_window} and Sec.~\ref{sec:results}.
Lastly, we follow the work by \cite{cai2025extrapolation} 
on length generalization and study whether distance generalization 
improves when models are trained on a main and an auxiliary task 
simultaneously, thereby allowing the model to infer 
from the auxiliary to the main task, or vice versa. We find that
distance generalization regularly benefits from transfer 
learning, but not always.

\subsubsection*{Our contributions}
We highlight the importance to thoroughly investigate 
distance generalization in transformers.
\begin{enumerate}
    \item 
    We find that performance improves in most cases when positional 
    schemes such as RoPE and ALiBi are turned off, which suggests
    that the role of positional encoding should be reconsidered. 
    \item We show that distance generalization can be used to 
    quantify an important, but hitherto little investigated question:
    How much does a model benefit when training on sets of
    increased data diversity? Our findings suggest that the
    answer is positive in absolute terms, but negative 
    when viewed relatively.
    \item We show that transfer learning \emph{can} positively impact distance
    generalization but also identify cases where transfer learning hinders
    generalization.
\end{enumerate}

%%%%%%%%%%%%%%%%%%%%%%%%%%%%%%%%%%%%%%%%%%%%%%%%%%%%
%%%%%%%%%%%%%%%%%%%%%%%%%%%%%%%%%%%%%%%%%%%%%%%%%%%%

\section{Related literature}
\subsubsection*{Length generalization.}
The ability of transformers to extrapolate from short 
training sequences to long test sequences, known as 
\emph{length generalization}, has been studied extensively
\cite{anil2022exploring,cho2024arithmetic,izzo2025quantitative,cai2025extrapolation,zhou2023algorithms}. Commonly,
artificial tasks such as copying, reverse copying or arithmetic 
tasks are used to assess length generalization capabilities \cite{kazemnejad2023impact,cai2025extrapolation,fan2024looped,xu2025principled}. In an effort to improve length generalization abilities, 
recent studies use several alternative approaches, 
including specialized positional encoding 
\cite{press2021train}, specific training protocols 
\cite{fan2024looped}, scratchpad prompting
\cite{anil2022exploring} and knowledge transfer from related tasks
\cite{cai2025extrapolation}.

\subsubsection*{Positional encoding.}
Following \cite{kazemnejad2023impact}, a strong impact of positional
encoding on distance generalization is expected. Many improvements to standard
learned and sinusoidal absolute positional encodings (APE)
\cite{vaswani2017attention} have been proposed. Currently, RoPE
\cite{su2024roformer} is the industry standard for large language models
and employed in many major models \cite{grattafiori2024llama,liu2024deepseek}. Despite RoPE's popularity, it offers poor length
generalization, spurring various works focused on 
mitigation of that shortcoming
\cite{kazemnejad2023impact,chen2023extending,peng2023yarn}. 
ALiBi, on the other hand, introduces a recency bias subtracted from 
the attention score which improves the ability for length generalization 
\cite{press2021train}.
In \cite{kazemnejad2023impact} the authors find that 
leaving out positional encoding
altogether is preferred for length generalization.

\subsubsection*{Transfer learning.}
Transfer learning refers to the transfer of knowledge 
from one task to a related task, it has been studied 
and applied extensively \cite{zhuang2020comprehensive}. 
In \cite{quirke2024understanding} the underlying mechanisms 
have been investigated for simple tasks. Namely, the identification
of computation circuits for addition and subtraction in transformers. 
The authors showed that a model trained on addition and subtraction 
and initialized with weights from a previously trained 
addition model can quickly achieve high accuracy on both tasks, 
which indicates knowledge reuse. 

There is somewhat less work with regard to transfer learning 
in the context of length generalization.
In \cite{cai2025extrapolation} the authors train a main task 
on a short context length together with a related auxiliary task,
the latter for longer context lengths. They find that transformers
are able to transfer knowledge from the auxiliary task to 
the main task, thus leading to improved length generalization 
for the main task.

\subsubsection*{Data diversity.}
In general, performance is strongly dependent on
training data selection procedures. With regard to the
ability of transformers to generalize, the authors of
\cite{zhou2024transformers} find that data formatting
can drastically change generalization outcomes. An
important determinant is training data diversity. In 
\cite{kawata2026shortcut} it is shown that 
increasingly diverse training data facilitates the 
formation of induction heads \cite{olsson2022context}, 
which enables the model in turn to generalize to 
out-of-distribution examples. Insufficient data diversity 
leads in contrast to positional shortcuts that are unable 
to generalize. In the context of length generalization, 
\cite{song2025out} shows that decreasing the size of the
training pool pushes the model into a memorization regime 
where generalization performance collapses.
In \cite{izzo2025quantitative} the authors compute a 
lower bound for the context length
required for length generalization to be possible.

\section{Experimental setup}\label{sec:experimental-setup}

Our dataset consists of token sequences of arbitrary length
from which a context window of length $T$ is randomly selected.
If not otherwise stated, we use $T = 256$. The token sequence
is equivalent to a concatenation of `tasks', akin to the
task switching framework presented in \cite{gros2025small}.
For our investigations, tasks correspond to one of the 
following three control tokens:

\begin{itemize}
    \item \textbf{Random task (\randct).} This is an auxiliary task
    that does --not-- contribute to the loss function. 
    The \randct control token signals the advent of randomly 
    generated tokens (random numbers), which are --not-- to be
    predicted. In general, using random tokens for copy tasks 
    prevents shortcut learning and ensures that models are 
    forced to copy tokens instead of relying on putative hidden patterns.

    \item\textbf{Delay copy task (\dcct).} For the delay copy
    task, control tokens appear in pairs:
     \begin{center}
    \includegraphics{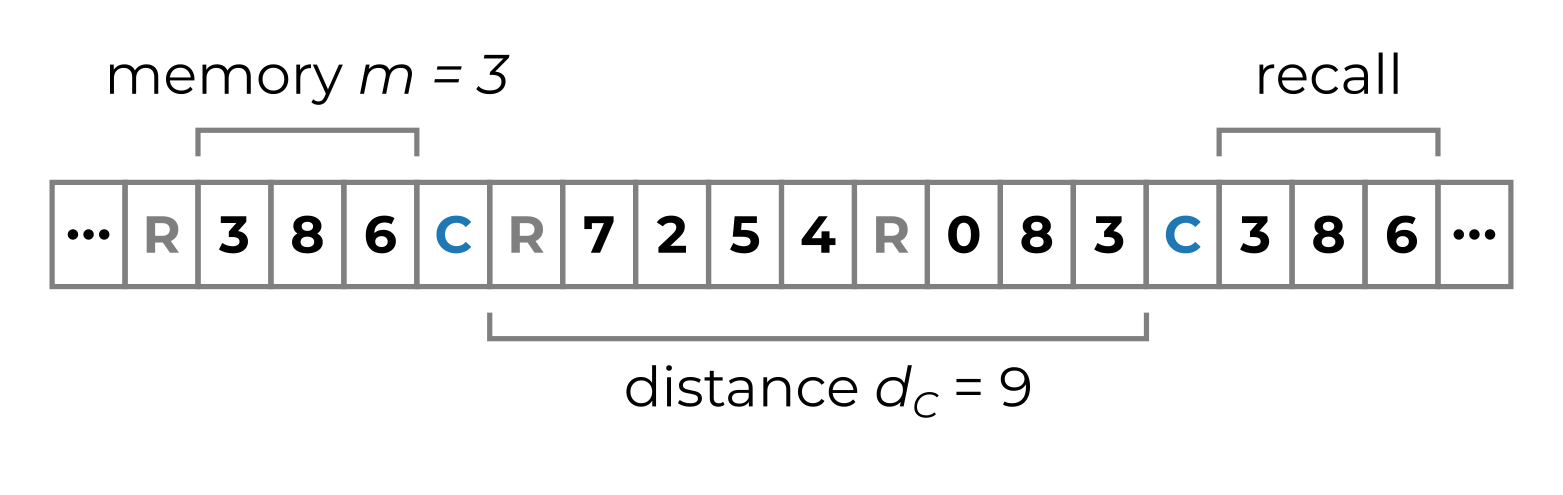}
    \end{center}
    The first
    \dcct token indicates that the $m\in\mathbb{N}$ preceding 
    tokens are to be copied with a delay at a later point in the 
    sequence, with the second \dcct token indicating the pasting 
    location. In between the pair of delay copy tokens, 
    other tasks may occur, in our case only the \randct task.
    The intra-pair \dcct-\dcct distance, denoted
    $d_C$, is sampled uniformly from a preset range, as detailed
    further below.

    \item\textbf{Selective delay copy task (\sdcct).} This task works
    equivalently to the delay copy task (\dcct). The \sdcct control 
    tokens also appear in pairs, where the first occurrence indicates 
    that the $m\in\mathbb{N}$ preceding tokens
    are to be copied selectively, with the second occurrence marking 
    the pasting location:
      \begin{center}
    \includegraphics{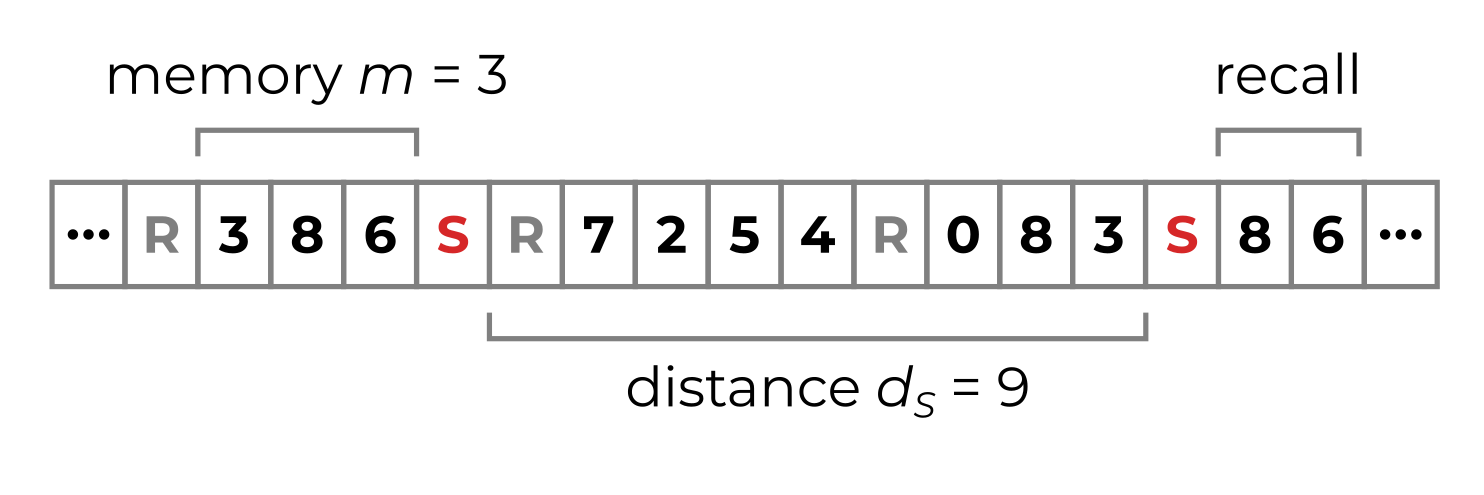}
    \end{center}
    In contrast to \dcct, not all tokens are to be copied, but only 
    tokens corresponding to even numbers. Importantly, the original order 
    of appearance needs to be maintained.
    In general, only a fraction of the source is hence to be recalled.
    In a dataset where both \dcct and \sdcct are present, the sampling
    ranges for the respective intra-pair distances $d_C$ and $d_S$ can
    be different.

\end{itemize}

\subsubsection*{Task switching framework.}
We train and evaluate on sequences of a fixed context length of
$T=256$, if not otherwise stated.  We consider three types
of datasets, $\dataset_{\randct \dcct}$,
$\dataset_{\randct \sdcct}$, and 
$\dataset_{\randct \dcct \sdcct}$, where the subscripts
indicate the set of tasks used. A given input sequence
pivots therefore between different synthetic tasks. As 
an example, here an extract of a sequence drawn from
$\dataset_{\randct \dcct \sdcct}$ with $N_\mathrm{base} = 10$:
\begin{center}
\includegraphics{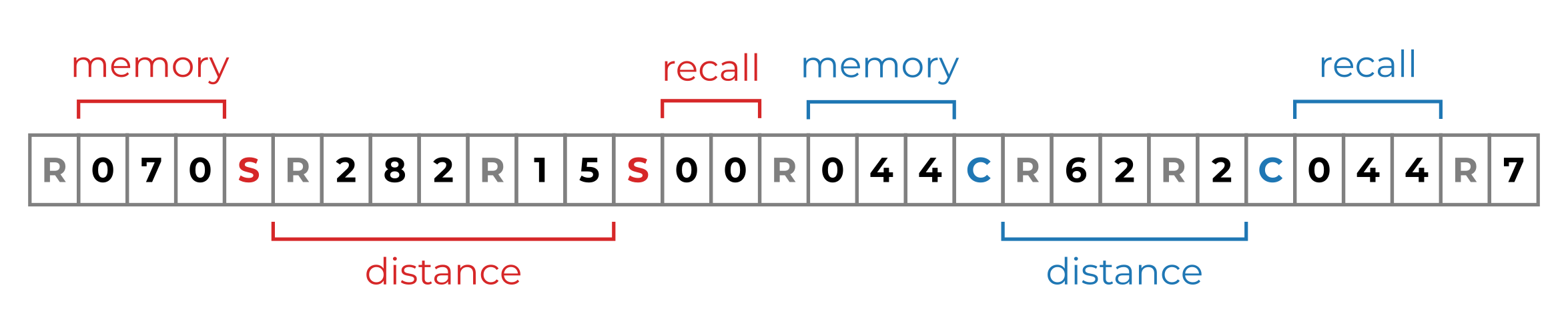}
\end{center}
For the task switching frequency a stochastic process is used, 
specifically, we sample task durations $\ell$ from a Binomial 
distribution,
\[
    P(\ell=k) = \binom{n}{k}p^k (1-p)^{n-k}\,,
\]
where $n=\ell_\mathrm{max}$ and $p = \mu / \ell_\mathrm{max}$.
We keep fixed parameter values $\mu = 6$ and $\ell_\mathrm{max} = 9$. All tasks
are executed on integer data tokens $t \in [0, \dots, N_\text{base})$ which
leads to a vocabulary size of $N_\text{base} + N_\text{tasks}$. In our
experiments we typically use $N_\text{base} = 128$.

\subsubsection*{Model.} For our experiments we use a conventional causal
decoder-only transformer with 8 layers and 8 attention heads and
model dimension 512. For a full specification of the architecture
please refer to App.~\ref{sec:app_expdetails}. 
We use the following positional encodings: ALiBi \cite{press2021train} and
RoPE \cite{su2024roformer}, which are tested against models
without positional encoding (NoPE). Details are given in
App.~\ref{sec:app_expdetails}.

\subsubsection*{Training and evaluation.}
Training is done using input-target pairs with a next-token 
prediction objective and teacher forcing, optimizing a
cross-entropy loss with an AdamW optimizer. For
a full overview of the hyperparameters refer to 
App.~\ref{sec:app_expdetails}. In the loss function we mask out 
random tasks (\randct) as well as the control
tokens, which are all unpredictable. 

Performance evaluation follows an equivalent protocol,
this time however for datasets with fixed distances
between source and recall, which cover both in-distribution
and out-of-distribution values.  Accuracies are 
computed over the full sequence where we assess correct
next-token predictions with teacher forcing. This means that 
we do not gauge performance under free-form multi-token 
generation, which is used at times in length generalization studies.
Our choice is motivated by the desire to avoid error compounding 
and to isolate recall performance in the delay copy tasks. 
Consistent with the training phase, random tasks and
control tokens are excluded in the accuracy metrics. We typically 
train on datasets containing a range of delay distances from which 
we randomly sample. At inference time, we evaluate performance
individually for delay distances of arbitrary size, probing in this 
way distance-generalization abilities.

\section{Results}\label{sec:results}

Conceptually, delay copy tasks could be solved using a range 
of generic machine-learning algorithms. For example, by storing 
source tokens into a dedicated cache, which could then be reused 
later on. This particular approach would be agnostic to delay distances, 
implying essentially perfect generalization. 

Transformers do not have access to dedicated memory units. Instead,
all positions within the context window are openly available. The 
problem transformers face is to identify the correct position of
source tokens. A possible solution would be: 
(a) to find the previous \dcct or \sdcct  token (or the respective \randct token), 
(b) to determine how many tokens have already been copied, and 
(c) to count back (or forth) in order to 
access the correct memory position. As discussed next, our results 
indicate that a bottleneck in this sequence of steps may be to resolve the
distance between the current and the preceding \dcct or \sdcct token.

%%%%%%%%%%%%%%%%%%%%%%%%%%%%%%%%%%%%%%%%%%%%%%%%
\begin{figure}[t]
    \centering
    \includegraphics[width=\textwidth]{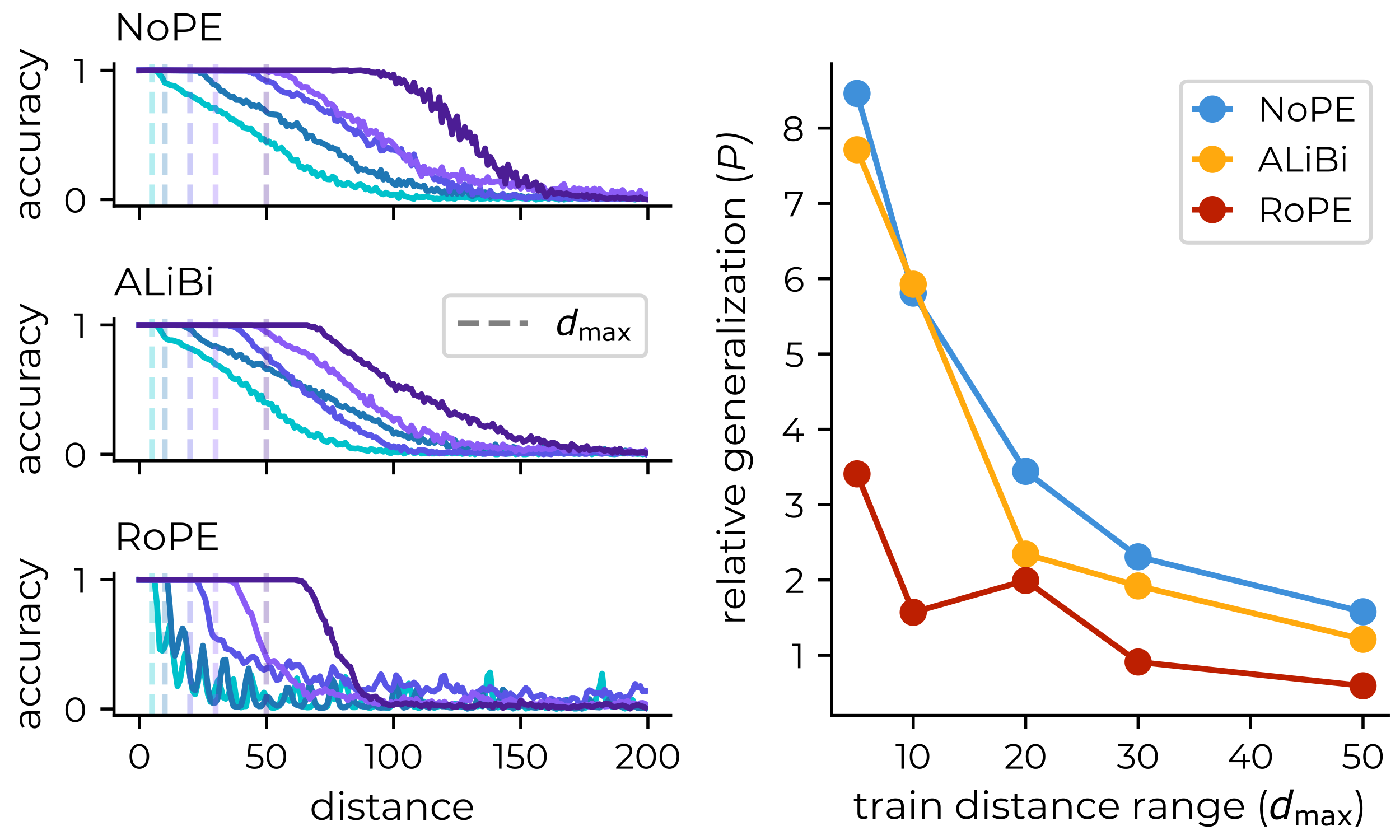}
  
    \caption{{\sl Influence of data diversity on distance generalization.} 
    Shown are results for $d_\mathrm{max} \in \{\textcolor[HTML]{00C2CB}{5}, 
    \textcolor[HTML]{1F77B4}{10}, \textcolor[HTML]{5955E6}{20}, 
    \textcolor[HTML]{8B5CF6}{30}, \textcolor[HTML]{4C1D95}{50}\}$,  
    where distances seen during training are in $[0,d_{\rm max}]$, compare~\eqref{eq_d_min_max}.
    Dashed lines in the left panels indicate the end of the training distance range, 
    i.e.\ $d_\mathrm{max}$, which is taken as proxy for data diversity. \\
    The right panel shows that diminishing returns are observed for
    relative distance generalization, as defined in \eqref{eq:rel_generalizaiton}.  
    This means that the absolute increase in distance generalization seen in
    the left panels with increasing data diversity is sub-linear with respect
    to data diversity $N_d=d_{\rm max}$.
    }
    \label{fig:dist_window}
\end{figure}
%%%%%%%%%%%%%%%%%%%%%%%%%%%%%%%%%%%%%%%%%%%%%%%%

\subsubsection{Positional encoding.}
In Fig.~\ref{fig:pes_by_task} we show results for the influence
of positional encoding on distance  generalization.
Transformers with three different position-encoding 
approaches have been trained, the two explicit encodings ALiBi 
and RoPE, and a model with no positional encoding (NoPE).
Training and evaluation are for datasets 
$\dataset_{\randct \dcct}$ 
(the basic copy task, Fig.~\ref{fig:pes_by_task} left) and $\dataset_{\randct \sdcct}$ 
(selective copy, Fig.~\ref{fig:pes_by_task} right).

During training, the distances between source and recall 
(the delays) are sampled uniformly between 
$d_\mathrm{min} = 15$ and $d_\mathrm{max} = 25$, as indicated by the 
gray-shaded region in Fig.~\ref{fig:pes_by_task}. 
The number $m$ of tokens to be copied is kept fixed,
at $m=10$.

All positional encoding schemes achieve high in-distribution
test accuracies, viz within the trained distance region. 
In agreement with previous studies on length generalization
\cite{chen2023extending,kazemnejad2023impact,peng2023yarn},
we observed that RoPE performs worst. Counter-intuitively,
NoPE out-performs both explicit positional encoding schemes,
which conforms however with similar observations for 
length generalization  \cite{kazemnejad2023impact}. This confirms
that causal transformers are able to dynamically learn to encode 
relative positions.

\subsubsection*{Training data diversity.}

In Fig.~\ref{fig:pes_by_task}, we kept both the number
of tokens to be copied fixed, $m=10$, as well as the
number $N_d$ of delay distances seen during training, 
namely at $N_d=10$ (which results from $10=25-15$). Next 
we study the impact of changing $N_d$, which we take as a proxy
for data diversity. For this we consider training datasets
for which recall distances are sampled uniformly between
$d_{\rm min}$ and $d_{\rm max}$, with
\begin{equation}
  d_\mathrm{min} = 0,\qquad\quad d_\mathrm{max}\in\{5, 10, 20, 30, 50\}\,,  
\label{eq_d_min_max}
\end{equation}
which implies $N_d=d_{\rm max}$. The results for
above set $\{d_{\rm max}\}$ used for our simulations are presented
in Fig.~\ref{fig:dist_window}. We use the
ratio $P$ between the cumulative out-of-distribution performance
$P_{\rm out}$ and the cumulative in-distribution
performance $P_{\rm in}$,
\begin{equation}\label{eq:rel_generalizaiton}
    P = \frac{P_{\rm out}}{P_{\rm in}} = 
    \frac{\sum_{d > d_{\mathrm{max}}} \mathrm{accuracy}(d)}{\sum_{d \leq d_{\mathrm{max}}}\mathrm{accuracy}(d)}\,,% \;\approx\; \frac{\int_{d > d_{\mathrm{max}}} \mathrm{acc}(d)\,\mathrm{d}d}{\int_{d \leq d_{\mathrm{max}}} \mathrm{acc}(d)\,\mathrm{d}d}\,,
\end{equation}
as a measure for assessing relative distance generalization
capabilities. Increasing data diversity leads to an increase
in $P_{\rm out}$, but to a decrease in $P$, as shown in
Fig.~\ref{fig:dist_window}. Given that $P_{\rm in} \approx N_d$,
this implies a sub-linear scaling of the
out-of-distribution performance $P_{\rm out}$
with data diversity $N_d = d_{\rm max}$. 
Diminishing returns are hence observed. 
This conclusion holds, because models reach 
an accuracy floor before the end of the evaluation range. 
Only in the opposite case
the denominator $P_\mathrm{in}$ in \eqref{eq:rel_generalizaiton} 
would lead to a structural bias.

\begin{figure}[t]
    \centering
    \includegraphics[width=\textwidth]{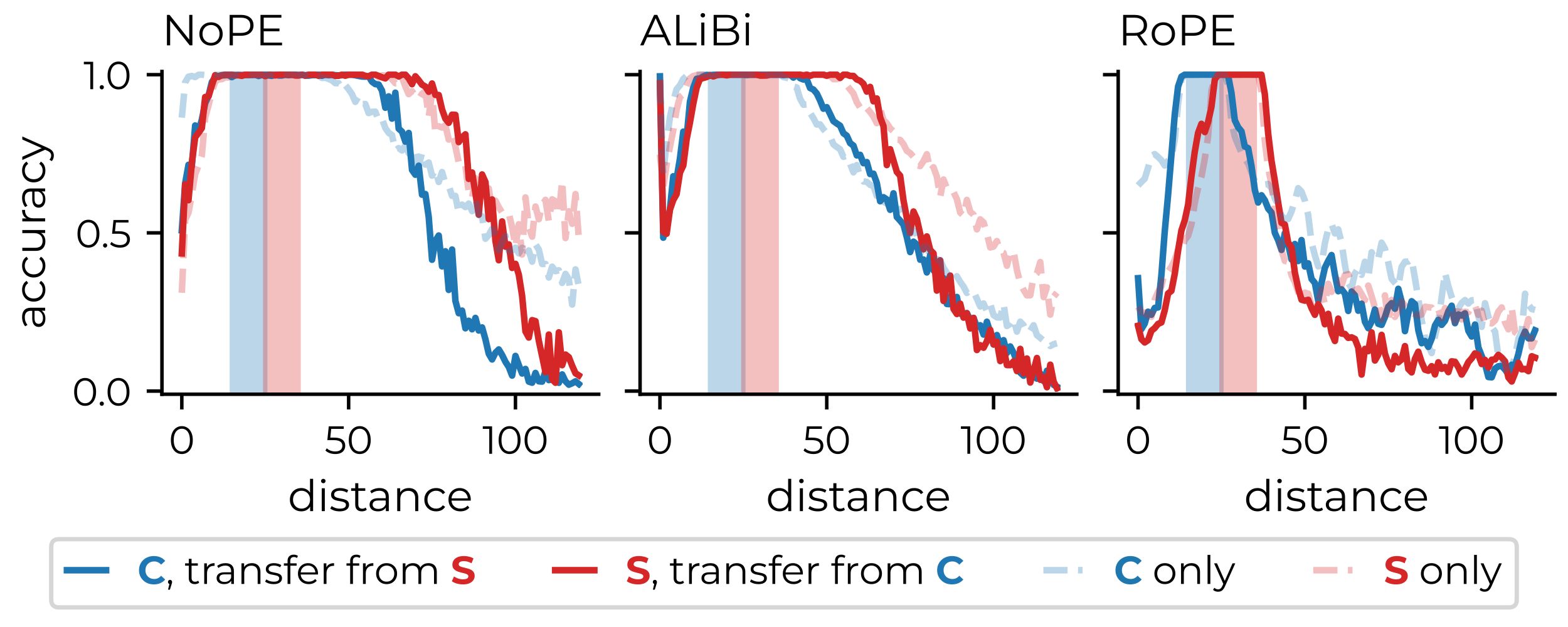}
    \caption{{\sl Transfer learning.} Shown are performance curves for three types of dataset:
    $\dataset_{\randct \dcct \sdcct}$ (solid lines), containing both the basic copy task (blue)
    and selective copy (red). Shaded regions indicate the range of the respective training
    delay distances.
    $\dataset_{\randct \dcct}$ (blue dashed), for comparison. Single task setting,
    here the basic copy task.
    $\dataset_{\randct \sdcct}$ (red dashed), equivalently for selective copy.
   }
    \label{fig:transfer_closeby}
\end{figure}

An interesting question is if there is a minimal 
data diversity, as suggested for the case of length
generalization \cite{izzo2025quantitative}, such that models 
may develop generalization capabilities only when trained
on datasets with larger diversity. As presented in
Fig.~\ref{fig:transfer_closeby}, our data does not
rule out this possibility. We leave detailed investigations
of this point for future studies.

%%%%%%%%%%%%%%%%%%%%%%%%%%%%%%%%%%%%%%%%%%%%%%%%
%%%%%%%%%%%%%%%%%%%%%%%%%%%%%%%%%%%%%%%%%%%%%%%%

\subsubsection*{Basic transfer learning.}
Using the $\dataset_{\randct \dcct \sdcct}$ dataset, for which
training sequences included both $\dcct$ and $\sdcct$ tasks,
we investigate inter-tasks performance effects. We are interested
in particular in the case that the supports for the sets of
training delay distances differ for $\dcct$ and $\sdcct$ tasks.

To be concrete, we define with $\{d_C\}$ and $\{d_S\}$ the sets
of delay distances present in $\dataset_{\randct \dcct \sdcct}$.
In Fig.~\ref{fig:transfer_closeby} we present results for 
$d_C\in[15,25]$ together with $d_S\in[25,35]$. Shown are four
performance curves:
\begin{itemize}
    \item $\dcct$ only. For comparison, results for the
    corresponding $\dataset_{\randct \dcct}$, viz when just
    as single task is present in the data and during testing.
    \item $\sdcct$ only. Correspondingly for selective copy.
    \item $\dcct$, with transfer from $\sdcct$. The performance
    of the basic copy task in the presence of selective copy.
     \item $\sdcct$, with transfer from $\dcct$. Correspondingly,
     the other way around.
\end{itemize}
For RoPE, mostly negative interference is observed.
Out-of-distribution performance curves for 
\dcct/\sdcct decay faster in the presence of
the other task (\sdcct, respectively \dcct). The second task
acts hence primarily as a distractor. This result
contrasts somewhat with \cite{cai2025extrapolation}, where 
a mostly positive impact of transfer learning on length 
generalization capabilities has been reported.

The situation is more differentiated for both ALiBi and NoPE.
One observes mostly positive transfer learning for moderate
out-of-distribution tests, but negative effects for increased
out-of-distribution delay distances.

%%%%%%%%%%%%%%%%%%%%%%%%%%%%%%%%%%%%%%%%%%%%%%%%
\begin{figure}[t]
    \centering
    \includegraphics[width=\textwidth]{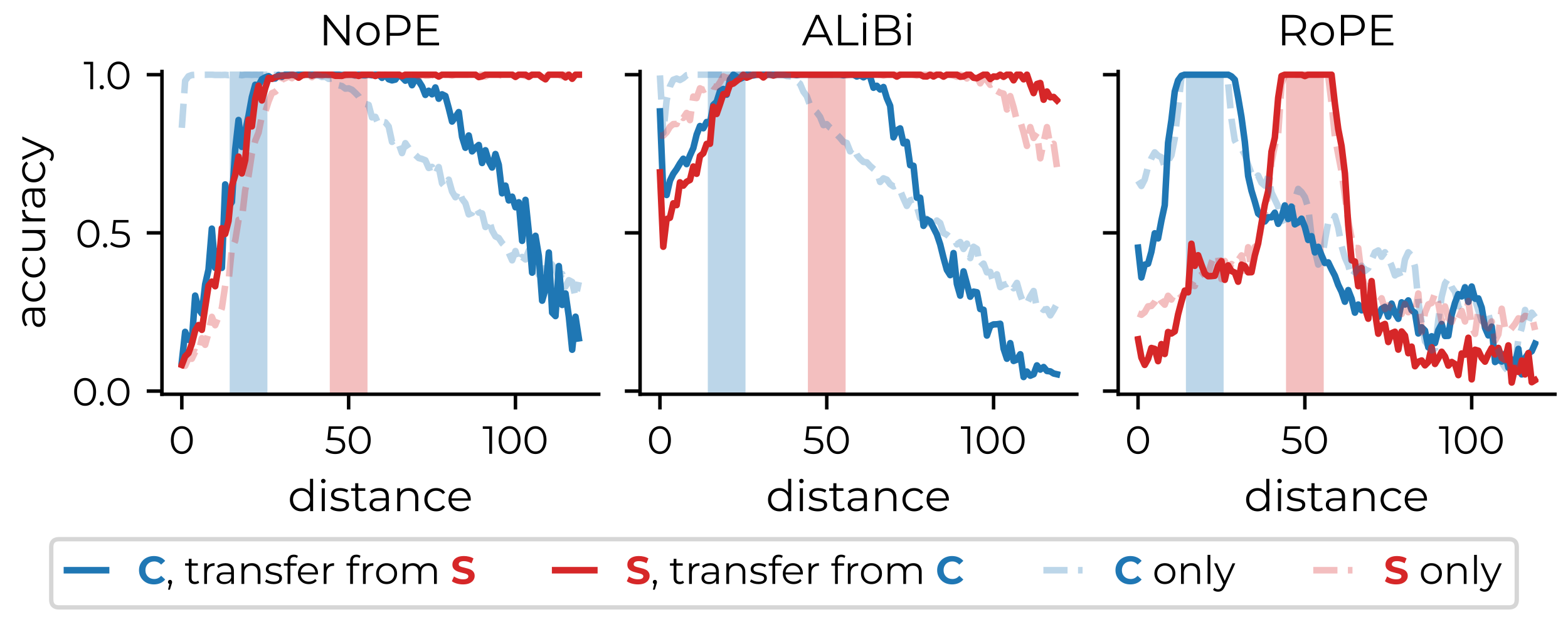}
    \caption{Transfer learning with enlarged task separation. For 
    three positional encoding schemes inference accuracies as a function of 
    the source-recall distance is shown, the procedure is otherwise identical to
    Fig.~\ref{fig:transfer_closeby}. The training distances are $[15,25]$ for
    the delay copy task \dcct, and $[45,55]$ for selective copy \sdcct. Solid lines
    are for transfer learning, when the model is trained on the joint dataset 
    $\dataset_{\randct\dcct\sdcct}$, dashed lines for datasets containing only one
    of the two tasks, either \dcct or \sdcct.}
    \label{fig:transfer_optimized_dists}
\end{figure}
%%%%%%%%%%%%%%%%%%%%%%%%%%%%%%%%%%%%%%%%%%%%%%%%

\subsubsection*{Complexity of transfer learning.}
In Fig.~\ref{fig:transfer_closeby} we did train the two tasks 
considered throughout this study on contiguous but disjoined 
distance ranges. As a natural next step, 
one may consider transfer learning in a setting where 
the two tasks are trained on well-separated distances ranges.
The corresponding results, as shown in Fig.~\ref{fig:transfer_optimized_dists},
demonstrate that transfer learning is a complex field. 
Here we will point to several central questions, leaving
an in-depth analysis to future work.

Firstly, for NoPE and ALiBi we observe an asymmetry in 
the direction of knowledge transfer, with constructive 
transfer learning occurring for large distances from 
\sdcct to \dcct. Compared to Fig.~\ref{fig:transfer_closeby}, 
one sees an improved generalization of
\dcct around the training distance
range of the \sdcct task. On the other hand, destructive
transfer learning effects are observed for both tasks 
around the training region of \dcct. Notably, 
the performance of the \dcct task itself is
negatively affected even within its own training region.
A possible explanation for the origin of the observed 
asymmetry for NoPE and ALiBi could be the higher structural 
complexity of \sdcct. If correct, this explanation would 
entail that transfer learning from a more complex to a 
less complex task might be favored. In our view, this
question deserves further attention.

Secondly, we observe significant differences between 
NoPE and ALiBi on one hand, and RoPE on the other hand.
Transfer learning is generally weak for RoPE, which has
however a key advantage: destructive effects
in the original training regions are not observed. 
For RoPE, distracting tasks seem to interfere less with 
bare training.

\section{Discussion}
In length generalization studies, poor extrapolation 
cannot be solely attributed to a failure of generalizing 
the task, as models are also facing out-of-distribution 
positions. To resolve this confounding, we propose to study 
distance generalization which tests models within a fixed 
context length, but with out-of-distribution token 
dependencies. In alignment with established practice \cite{cai2025extrapolation,kazemnejad2023impact}, we 
consider synthetic tasks, in particular two types of copy 
task with delay, involving respectively full and selective 
copying.

Our results suggest that distance generalization is 
highly dependent on the type of positional encoding used, cf.~Fig.~\ref{fig:pes_by_task}. We find that RoPE
falls behind other types of positional encoding, 
including models without any explicit positional 
encoding (NoPE), in agreement with length-generalization
studies \cite{kazemnejad2023impact}. 
For distance generalization, the effect is particularly striking, since RoPE is not facing 
untrained angular regimes at test time. As we show 
in Fig.~\ref{fig:app_model_size_depend}, explicit 
positional encoding is actually necessary for small 
models, yet further investigations are required in 
this regard. Follow-up studies may investigate more 
diverse tasks, together with in-depth circuit and attention 
analyses. Ultimately distance generalization needs to be
tested in pretrained models.

A number of concerns may be raised with
regard to the experimental setup used.
For transfer learning, models are trained 
on joint datasets $\dataset_{\randct\dcct\sdcct}$ 
without altering task-frequency parameters.
Training on $\dataset_{\randct \dcct \sdcct}$ will
result in fewer task-specific examples than in a 
single-task setting which could raise concerns about 
the origins of the effects reported 
in Fig.~\ref{fig:transfer_closeby} and \ref{fig:transfer_optimized_dists}. 
We note, however, that the in-distribution accuracy 
remains saturated in the joint setting shown in
Fig.~\ref{fig:transfer_closeby} and that
the destructive effect visible in
Fig.~\ref{fig:transfer_optimized_dists} 
appears within the training range of
\dcct. An exposure deficit alone would be expected 
to show up as a uniform degradation rather than
specific to certain distances. 

Second, following work on length generalization (e.g.~\cite{cai2025extrapolation}), the best of five 
runs ranked by cumulative accuracy have been used. 
This choice is deliberate: our questions concern 
whether a given architecture can resolve unseen 
distances at all. Seeds that fail to learn the 
copy mechanism in the first place are uninformative about 
distance generalization. As shown in Appendix B,
differences between different runs are actually 
minor.

% \printbibliography
\bibliographystyle{splncs04}
\bibliography{notes}

\appendix
\section{Experimental Details}
\label{sec:app_expdetails}
\subsection{Model Architecture}
We train a standard causal decoder-only Transformer model, following common design choices in prior work such as \cite{vaswani2017attention}. The model consists of 8 Transformer layers with a hidden size of 512 and 8 attention heads (head dimension 64). Each layer contains a feedforward MLP with ReLU activation and dropout applied with rate 0.1.
We test three common positional embeddings, namely RoPE \cite{su2024roformer}, ALiBi \cite{press2021train} and a model with no positional embedding (NoPE). Input and output embeddings are untied. Model details are summarized in Tab.~\ref{tab:model}.

\subsection{Training Procedure}
The model is trained from scratch using the AdamW optimizer \cite{kingma2014adam,loshchilov2017decoupled}. We use a learning rate of $3\times10^{-4}$ with linear warm-up over the first 3{,}000 steps, followed by a constant schedule. Weight decay is set to 0.05. Training is performed with a batch size of 64 for 40{,}000 update steps. The training objective is standard autoregressive next-token prediction. To avoid perplexity early on in the sequence, we exclude the first few tokens in the sequence from the loss and accuracy computation. Different values did not change results. We settled for a very conservative default value of 18 (which is 2$\times$ the default maximum task length). Training details are summarized in Tab.~\ref{tab:training}.

\begin{table}[h]
\centering
\begin{minipage}[t]{0.48\textwidth}
\centering
\caption{Model architecture configuration.}
\label{tab:model}
\begin{tabular}{lc}
\toprule
Parameter & Value \\
\midrule
Number of layers & 8 \\
Hidden size & 512 \\
Attention heads & 8 \\
Head dimension & 64 \\
Dropout & 0.1 \\
MLP activation & ReLU \\
\bottomrule
\end{tabular}
\end{minipage}
\hfill
\begin{minipage}[t]{0.48\textwidth}
\centering
\caption{Training configuration.}
\label{tab:training}
\begin{tabular}{lc}
\toprule
Parameter & Value \\
\midrule
Optimizer & AdamW \\
Learning rate & $3\times10^{-4}$ \\
Weight decay & 0.05 \\
Warmup steps & 3{,}000 \\
Batch size & 64 \\
Training steps & 40{,}000 \\
Loss and accuracy mask until & position 18 \\
\bottomrule
\end{tabular}
\end{minipage}
\end{table}

\subsection{Dataset}
We train on a synthetic multi-task sequence dataset in accordance with the task switching framework with three
different tasks: random (\randct), delay copy (\dcct) and selective delay copy (\sdcct). The tasks are detailed in 
the main text, see Sec.~\ref{sec:experimental-setup}. 
Dataset details are summarized in Tab.~\ref{tab:dataset}

\begin{table}[h]
\centering
\caption{Dataset configuration.}
\label{tab:dataset}
\begin{tabular}{lc}
\toprule
Parameter & Value \\
\midrule
Number of tokens & 10.000.000 \\
Sequence length & 256 \\
\# distinct data tokens & 127 \\
Memory for copying tasks & 10 \\
Training distance for copying tasks & $\sim \operatorname{unif}(d_\mathrm{min}, d_\mathrm{max})$ \\
\bottomrule
\end{tabular}
\end{table}

\section{Additional Results}
\label{sec:app_add_results}

%%%%%%%%%%%%%%%%%%%%%%%%%%%%%%%%%%%%%%%%%%%%%%%%
\begin{figure}[ht]
    \centering
    \includegraphics[width=\textwidth]{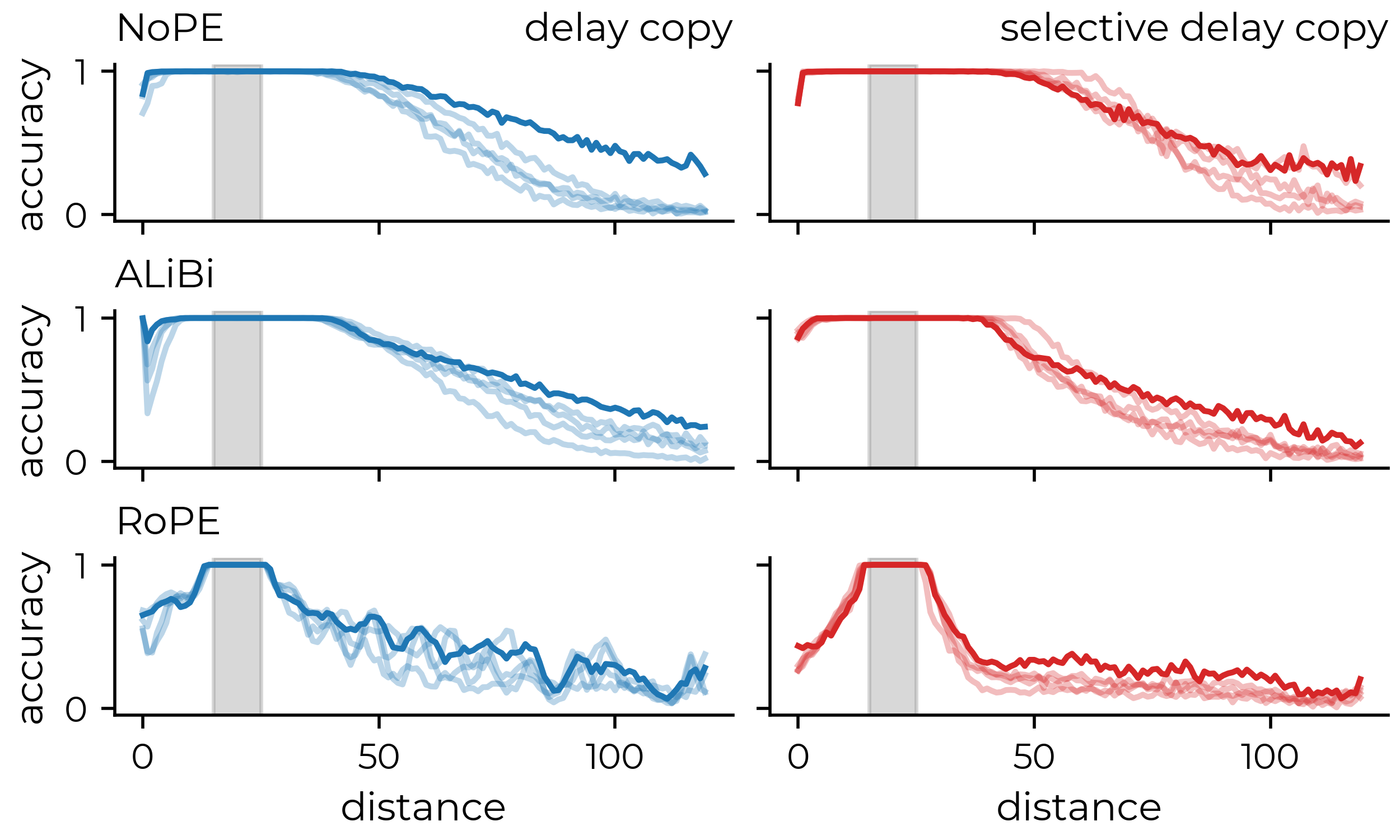}
    \caption{For different positional encodings, we show different training runs for (left) the delay copy task (\dcct),
    $\dataset_{\randct \dcct}$, and (right) the selective delay copy task (\sdcct), $\dataset_{\randct \sdcct}$,
    where $d_{C/S}\sim\operatorname{unif}(15, 25)$.
    In solid lines we show the best performing runs, measured by the integral over the full plotted distance range. These models
    are presented in the main text results.}
    \label{fig:different_runs}
\end{figure}
%%%%%%%%%%%%%%%%%%%%%%%%%%%%%%%%%%%%%%%%%%%%%%%%
\subsubsection*{Variance of training runs.}
In the main text results we generally present the best-performing model out of five unique training runs with equal parameters,
where we measure the performance by the cumulative accuracy
\begin{equation}\label{eq:appx_performace}
    \mathrm{performance} = \sum_{d=1}^{120} \mathrm{accuracy}(d)\,.
\end{equation}
Fig.~\ref{fig:different_runs} presents a variety of training runs for different positional encodings and different tasks, complementing
main-text Fig.~\ref{fig:pes_by_task}.
For other main-text results we use the same best-model policy.

\subsubsection*{The impact of the model size on distance generalization.}
\begin{figure}[t]
    \centering
    \includegraphics[width=\textwidth]{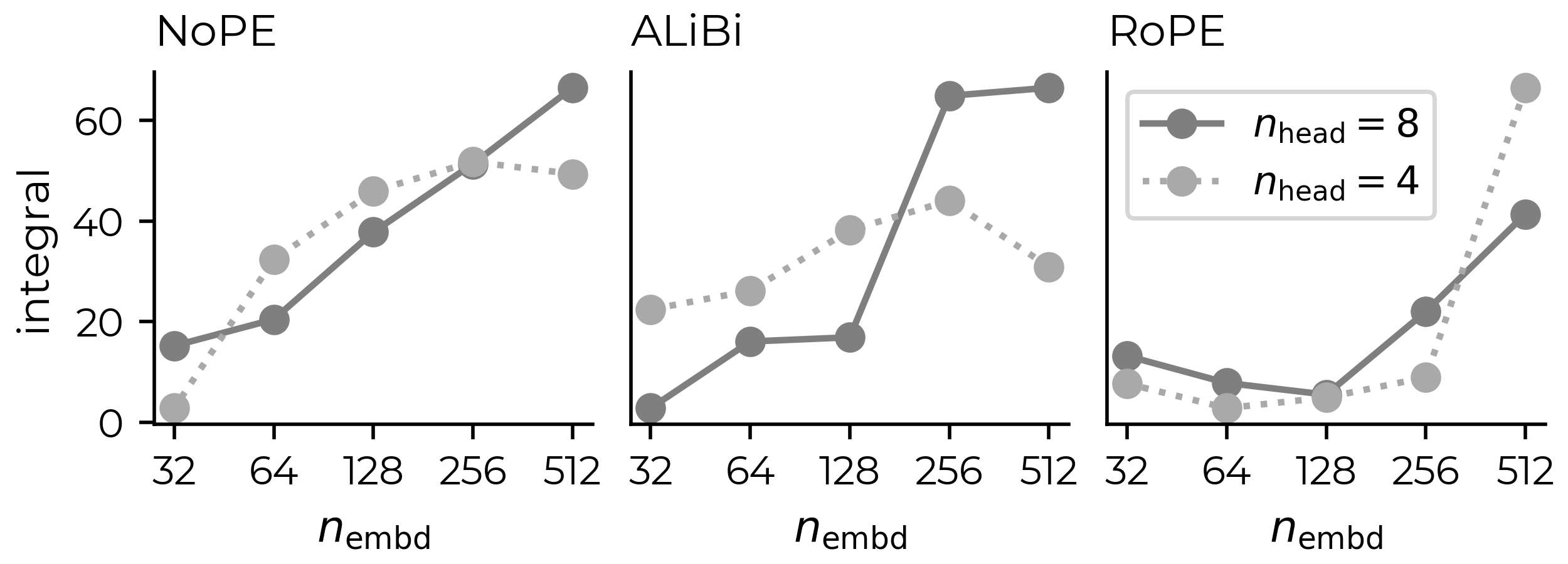}
    \caption{While we show that NoPE often performs well at distance generalization, we observe that it falls behind
    explicit encodings, ALiBi and RoPE, when the model size is small. For different positional encodings, we test and evaluate on
    a dataset $\dataset_{\randct \dcct}$, where training distances $d_C$ of the delay copy task (\dcct) are sampled from 
    $d_C \sim \operatorname{unif}(0, 10)$ (gray shaded region).
    For each positional encoding we test different model sizes. The number of layers is kept at 8 while we
    vary head size and model dimension. We show the cumulative
    accuracy for different model configurations \eqref{eq:appx_performace}.}
    \label{fig:app_model_size_depend}
\end{figure}

For the main text results we considered a model with embedding dimension $n_\mathrm{embd} = 512$, 8 heads and 8 layers.
In Fig.~\ref{fig:app_model_size_depend} we present results for 4 and 8 heads with
$n_\mathrm{embd} \in [2^i \colon i=5,\dots,9]$. Keeping the number of layers fixed at 8, as well as keeping all other 
model details as presented in App.~\ref{sec:app_expdetails}. We observe that distance generalization ability is coupled
to model size, where larger models tend to generalize better across the board. For ALiBi in a model with 8 heads
we find that smaller models break down significantly faster than their larger counterparts. We hypothesize that the
increased number of ALiBi-slopes can distract the model if parameter counts are not sufficient.

%%%%%%%%%%%%%%%%%%%%%%%%%%%%%%%%%%%%%%%%%%%%%%%%
%%%%%%%%%%%%%%%%%%%%%%%%%%%%%%%%%%%%%%%%%%%%%%%%

\subsubsection*{NoPE, albeit offering good distance generalization, breaks down in small models.}
In the main text we show, supporting findings for length generalization by \cite{kazemnejad2023impact}, that
NoPE offers in many situations outstanding distance generalization performance. An important caveat is, however, 
that NoPE requires a certain model size to encode positions effectively. For small-scale models, NoPE even struggles
to learn the mechanism of the delay copy task (\dcct) within distribution, see Fig.~\ref{fig:app_model_size_depend},
while explicit positional encodings are much more stable for smaller models.

\end{document}